# A Randomized Approximation Analysis of Logic Sampling

R. Martin Chavez and Gregory F. Cooper

chavez@sumex-aim.stanford.edu
cooper@sumex-aim.stanford.edu
Section on Medical Informatics
Medical School Office Building, X-215
Stanford University
Stanford, California 94305-5479

In recent years, researchers in decision analysis and artificial intelligence (AI) have used Bayesian belief networks to build models of expert opinion. Using standard methods drawn from the theory of computational complexity, workers in the field have shown that the problem of exact probabilistic inference on belief networks almost certainly requires exponential computation in the worst case [3]. We have previously described a randomized approximation scheme, called BN-RAS, for computation on belief networks [1, 2, 4]. We gave precise analytic bounds on the convergence of BN-RAS and showed how to trade running time for accuracy in the evaluation of posterior marginal probabilities. We now extend our previous results and demonstrate the generality of our framework by applying similar mathematical techniques to the analysis of convergence for logic sampling [7], an alternative simulation algorithm for probabilistic inference.

## 1.0 Introduction

Given truth assignments for a set $E$ of random variables in a belief network, an algorithm for Probabilistic Inference in Belief NETworks (PIBNET) computes the posterior probabilities for the outcomes of a specified node $X$. PIBNET is hard for **NP**, by reduction from 3-satisfiability in the propositional calculus [3]. That classification has focused research on approximate methods, special-case techniques, heuristics, and analyses of average-case behavior.

There now exists a number of algorithms for exact probabilistic inference in belief networks: the message-passing algorithm of Pearl [12], the triangulation method of Lauritzen and Spiegelhalter [10], and others. Previous approximation algorithms include the Markov-simulation scheme of Pearl [13, 14], Henrion's logic sampling [7], and the randomized approximation scheme (ras), known as BN-RAS, which we have previously demonstrated [1]. Heckerman has proposed a special-case algorithm for certain kinds of two-level belief networks [6]. Each algorithm has computational properties that render it attractive for inference on certain kinds of networks. The NP-hard classification suggests, however, that no algorithm can provide a definitive efficient solution for all inference problems.

A *randomized approximation scheme (ras)* is, by definition, an algorithm that computes, with low probability of

Methods and Procedures

failure, a result that lies arbitrarily close to the true answer [8, 9]. Moreover, a ras must terminate within time that is a linear function of the reciprocal of the error and the reciprocal of the failure probability. We can define approximation schemes that guarantee either interval or relative error bounds, with high probability.

BN-RAS derives from Markov-simulation algorithms originally proposed by Pearl. We now introduce a new ras, BN-RAS-LS, based on logic sampling. We do not modify the original algorithm for logic sampling; rather, we specify a convergence analysis that transforms the logic-sampling method into a ras.

## 2.0 Methods and Procedures

Suppose that we wish to compute all posterior probabilities in a belief network to within an interval error $\alpha$. Suppose, in addition, that we are willing to tolerate a small probability $1 - \delta$ that the algorithm fails to converge within the $\alpha$ bound. Let $\mu$ represent the true posterior marginal probability of the node under consideration; $\hat{Y}$ denotes the approximate probability computed by the algorithm. The $(\alpha, \delta)$ convergence criterion may be written

$$P(|\hat{Y} - \mu| \leq \alpha) \geq \delta. \qquad (1)$$

The detailed argument, based on Chebyshev's inequality [9], reveals that

$$N \geq \frac{1}{4(1-\delta)\alpha^2} \qquad (2)$$

guarantees the $(\alpha, \delta)$ convergence criterion, where $N$ is the total number of simulation trials. Each trial corresponds to the choice of a joint instantiation, consistent with known evidence, for all the nodes in the belief network. Inasmuch as $N$ is a polynomial in $1/\delta$ and $1/\alpha$, our sampling scheme with its convergence criterion defines a ras for PIBNET.

An algorithm for logic sampling simulates a value for every variable in the model, in graphical order, at each trial. When a variable $X$ is simulated, the values of its conditioning variables have already been selected; logic sampling chooses a new value for $X$ based on its conditional-probability matrix. The forward-propagation technique of logic sampling suffers as the size of the set of findings $F$ grows. Trial samples that are consistent with $F$ are called *successful trials*. Samples that do not correspond to the findings in $F$ must be discarded, and the algorithm's performance deteriorates as a result. (See [7] for additional details.)

We view logic sampling over a belief network with a set of findings $F$ to be a Bernoulli process with a binomial distribution

$$P(K > N) = 1 - \sum_{i=0}^{N} \binom{n}{i} p^i q^{n-i}, \qquad (3)$$

where $K$ is the number of successful logic-sampling trials, $N = 1/[4(1-\delta)\alpha^2]$ is the number of trials needed to ensure $(\alpha, \delta)$ convergence, $n$ is the total number of samples (including successes and failures), $p = P(F)$ is the prior probability of the set of findings (i.e., the probability of a successful trial), and $q = 1 - p$.

If $K > N$, then we have performed more than $N$ successful trials, and therefore $P(|\hat{Y} - \mu| \leq \alpha) \geq \delta$. Thus, if $P(K > N) \geq \sigma$, then by substitution of $P(|\hat{Y} - \mu| \leq \alpha) \geq \delta$ for $K > N$, we obtain that $P(P(|\hat{Y} - \mu| \leq \alpha) \geq \delta) \geq \sigma$.

We now introduce an intermediate result:

**LEMMA 1.**

Let $I = |P'(z|e) - P(z|e)|$, where $P'(z|e)$ is an unbiased estimator of $P(z|e)$, and where $z$ and $e$ are random variables. If $P(P(I \leq b) \geq c) > d$, then $P(I \leq b) > cd$, for arbitrary constants $b$, $c$, and $d$.

**Proof.** First, by Markov's inequality:

$$P(X \geq c) < \frac{E(X)}{c}. \qquad (4)$$

Let $X = P(I \leq b)$. Then, by applying (4) to the antecedent of the theorem, we have that $P(X \geq c) \leq ((E(X))/c)$. Note that the theorem also specifies that $d < P(X \geq c)$. Combining, we see that $d < P(X \geq c) \leq (E(X))/c$. Thus, $E(X) > cd$. Now, $E(X) = X = P(I \leq b)$, so $P(I \leq b) > cd$ and the lemma is proved. Q.E.D.

If $P(K>N) \geq \sigma$, then $P(P(|\hat{Y}-\mu| \leq \alpha) \geq \delta) \geq \sigma$ and thus by Lemma 1:

$$P(|\hat{Y}-\mu| \leq \alpha) \geq \delta\sigma. \quad (5)$$

Now define a function

$$f(n, p, N) = \sum_{i=0}^{N} \binom{n}{i} p^i q^{n-i}. \quad (6)$$

In addition, let

$$g(\sigma, p, N) = \min\{n: 1-f(n, p, N) \geq \sigma\}. \quad (7)$$

If we perform $g(\sigma, p, N)$ trials, then $P(K>N) \geq \sigma$ and relation 5 is satisfied. We therefore require an efficient algorithm for computing an upper bound on $g(\sigma, p, N)$.

Consider the following normal approximation to the binomial distribution $f(n, p, N)$:

$$\hat{f}(n, p, N) = \Phi\left(\frac{N-np+\frac{1}{2}}{\sqrt{npq}}\right) - \Phi\left(\frac{-np-\frac{1}{2}}{\sqrt{npq}}\right), \quad (8)$$

where $\Phi$ is the standard normal distribution. In [11, page 170], it is shown that

$$|\hat{f}(n, p, N) - f(n, p, N)| \leq \frac{0.14}{\sqrt{npq}}. \quad (9)$$

Therefore,

$$g(\sigma, p, N) \leq \min\{n: \hat{f}(n, p, N) \leq 1-\sigma-\frac{0.14}{\sqrt{npq}}\}. \quad (10)$$

Now consider the approximation

$$\Phi'(x) = 1 - 0.5[1.0 + (0.115194)x^2 + 0.000344x^3 - 0.019527x^4]^{-4}, \quad (11)$$

which carries a maximum interval error of $2.5 \times 10^{-4}$ [11, page 51]. (We give approximations, with interval errors, because tables are not available for the wide range of values that we may require.) If we let

$$f'(n, p, N) = \Phi'\left(\frac{N-np+\frac{1}{2}}{\sqrt{npq}}\right) - \Phi'\left(\frac{-np-\frac{1}{2}}{\sqrt{npq}}\right), \quad (12)$$

then $|\hat{f}(n, p, N) - f'(n, p, N)| \leq 5.0 \times 10^{-4}$, and we have

$$g(\sigma, p, N) \leq \min\{n: f'(n, p, N) \leq 1-\sigma-\frac{0.14}{\sqrt{npq}}-5.0 \times 10^{-4}\}. \quad (13)$$

To find an appropriate value for $n$, we perform a binary search in the range $[N, N^2/p]$ and look for values that satisfy the inequality in (13). (The approximation $f'(n, p, N)$ is a monotonically decreasing function on $n$.) Clearly, $n \geq N$, because we require at least $N$ successful trials. The choice of $N^2/p$ is purely arbitrary. Notice that as long as

$$\sigma \leq 0.99954 - \frac{0.14}{N\sqrt{q}} - f'(\frac{N^2}{p}, p, N), \quad (14)$$

$n = N^2/p$ will satisfy (13) and will therefore ensure that $P(K>N) \geq \sigma$. Hence, for most $\sigma$ of interest, we have given a binary-search algorithm that yields an upper bound $g_u(\sigma, p, N)$ on $g(\sigma, p, N)$. We can accommodate $\sigma$ closer to 1 by choosing finer approximations to the normal distribution, or by expanding the upper end of the search interval beyond $N^2/p$.

Note that, in general, $p = P(F)$ is not known exactly. We can, however, calculate a lower bound $p'$ on $p$ as follows:

$$p \geq p' = \prod_{X \in F} \min_{\pi_X}[P(X|\pi_X)], \quad (15)$$

where the $X$ are nodes in the finding set $F$, and the $\pi_X$ denote the parents of $X$. If we perform $g_u(\sigma, p', N)$ trials, then $P(K > N) \geq \sigma$. Poor lower bounds will, of course, cause the performance of the algorithm to deteriorate.

Our arguments prove the following theorem:

**THEOREM 1.**

To obtain $P(|\hat{Y} - \mu| \leq \alpha) \geq \delta\sigma = \omega$ for logic sampling, with $\delta$ and $\sigma$ such that $\sigma$ satisfies (14), perform $g_u(\sigma, p', N)$ trials using the standard protocol for logic sampling, and score the outcomes in the usual fashion.

As trials are scored, we can incrementally reapply a similar analysis to compute a new value for $g_u(\sigma, p', N)$ in light of the trial successes already encountered. Moreover, it is also possible to compute a lower bound $g_l(\sigma, p, N)$ on $g(\sigma, p, N)$, derived from the relation

$$g(\sigma, p, N) \geq \min\left\{n: f'(n, p, N) \leq 1 - \sigma + \frac{0.14}{\sqrt{npq}} + 5.0 \times 10^{-4}\right\}. \quad (16)$$

Also,

$$p \leq p'' = \prod_{X \in F} \max_{\pi_X} [P(X | \pi_X)]. \quad (17)$$

Initially, we expect to have to perform at least $g_l(\sigma, p'', N)$ trials in order to achieve the convergence criterion. Knowing the lower bound can be very useful in indicating whether simulation for a given set of parameters is worth doing. Note that estimate-based approaches to error analysis [7] cannot yield such a priori lower bounds, because those approaches require several successful trials; that requirement, in turn, defeats the objective of avoiding simulation in cases where successful trials are rare.

Other workers in the field have previously given procedures for estimating the accuracy of logic-sampling estimates. In particular, Henrion [7] uses a sample to compute an estimate of the standard error of a probability of interest. Our approach differs in that we use Chebyshev's and Markov's inequalities, as applied to the probabilities of the belief network, to specify *a priori* upper and lower bounds on the expected number of trials required for convergence. By ap-proximating the binomial distribution, we avoid making assumptions about asymptotic behavior, such as convergence to the normal distribution. Without such an analysis, a method for computing probability estimates cannot be classified as a ras.

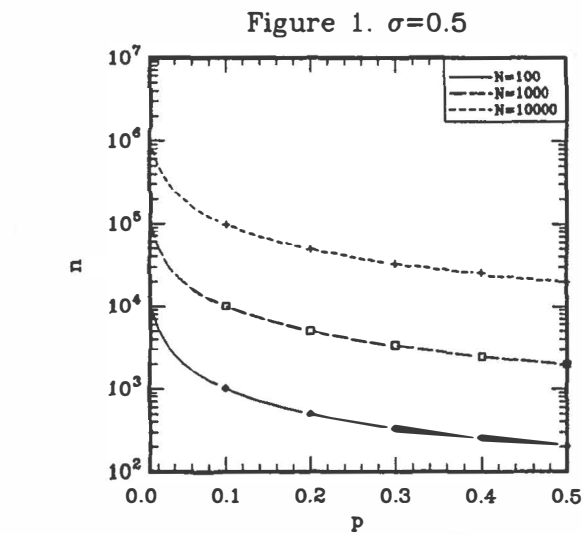

Figure 1. $\sigma = 0.5$

Figure 1 plots the number of trials suggested by our methodology, for various values of the probability of the set of findings and for several values of $N$, with $\sigma$ held to 0.5. Figures 2 and 3 plot corresponding values for $\sigma = 0.9$ and $\sigma = 0.99$, respectively. On log-linear scales, the value of $\sigma$ makes little difference; the probability $p$, on the other hand, strongly determines the expected number of trials.

## 3.0 Discussion

We have shown how to compute a priori bounds on the randomized computation of marginal posterior probabilities in belief networks. We have applied our analytic techniques both to Markov simulation [1] and, in this paper, to logic sampling. To characterize the performance of logic sampling in analytic terms, we have employed the area-estimation technique of Karp and Luby, Chebyshev's inequality, Markov's inequality, and normal approxima-tions to the binomial distribution. The resulting ras speci- fies the expected number of trials needed for convergence as a function of the interval error, the failure probability, and the prior probability of the evidence set.

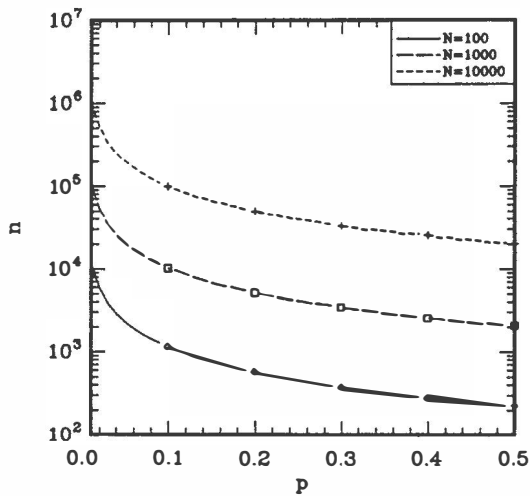

Figure 2. $\sigma=0.9$

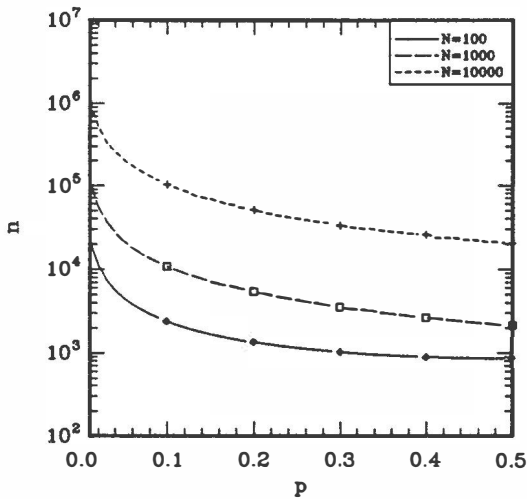

Figure 3. $\sigma=0.99$

The number of trials specified by $g_u(n, p', N)$ is a *sufficient* number of trials, such that $P(|\hat{Y}-\mu| \leq \alpha) \geq \omega$ holds true *at the start*, before we begin simulation, for every posterior probability $\mu$ in the network. After we perform $g_u(n, p', N)$ trials, however, we are not guaranteed that $P(|\hat{Y}-\mu| \leq \alpha) \geq \omega$ still holds, because it is possible that $K$ is less than $N$. Note, on the other hand, that we gain information about convergence from the number of successful trials as the simulation proceeds. We view that adaptive behavior, which distinguishes BN-RAS-LS from its Markov-simulation predecessor BN-RAS, as one of the algorithm's most appealing properties.

The analysis of running time has important practical consequences. The selection of an appropriate algorithm for probabilistic inference depends crucially on the parameters of the problem. A meta-algorithm that selects inference techniques can apply those stochastic algorithms with the best expected convergence for a given inference task.

The techniques we have described do not apply directly to modified stochastic algorithms for probabilistic inference, including importance sampling. We expect that future research will illuminate the running time of those more sophisticated algorithms, and thereby further facilitate the tailoring of belief-network inference techniques to knowledge bases for real-world applications.

## 4.0 Acknowledgments

Lyn Dupré edited the manuscript.

This work has been supported by grant IRI-8703710 from the National Science Foundation, grant P-25514-EL from the U.S. Army Research Office, the Medical Scientist Training Program under grant GM07365 from the National Institutes of Health, and grant LM-07033 from the National Library of Medicine. Computer facilities were provided by the SUMEX-AIM resource under grant LM-05208 from the National Institutes of Health.

## 5.0 References

1. R. M. Chavez and G. F. Cooper. A randomized approximation scheme for the Bayesian inferencing problem. Technical Report KSL-88-72, Knowledge Systems Laboratory, Stanford University, Stanford, CA, April 1989. To appear in *Networks*.

2. R. M. Chavez and G. F. Cooper. An empirical evaluation of a randomized algorithm for probabilistic inference. In *Proceedings of the Fifth Workshop on Uncertainty in Artificial Intelligence,* Windsor, Ontario, August 1989, pages 60-70.

3. G. F. Cooper. The computational complexity of probabilistic inference using Bayesian belief networks. *Artificial Intelligence* 42 (1990), pages 393-405.

4. R. M. Chavez. Hypermedia and randomized algorithms for medical expert systems. In *Proceedings of the Thirteenth Symposium on Computer Applications in Medicine*, Washington, DC, November 1989. To appear in *Computer Methods and Programs in Biomedicine*.

5. M. R. Garey an D. S Johnson. *Computers and Intractability: A Guide to the Theory of NP-Completeness*. W. H. Freeman and Company, New York, NY, 1979.

6. D. E. Heckerman. A tractable inference algorithm for diagnosing multiple diseases. In *Proceedings of the Fifth Workshop on Uncertainty in Artificial Intelligence,* Windsor, Ontario, August 1989, pages 174-180.

7. M. Henrion. Propagating uncertainty in Bayesian networks by probabilistic logic sampling. In J. F. Lemmer and L. N. Kanal, eds., *Uncertainty in Artificial Intelligence 2*, North-Holland, 1988, pages 149-163.

8. M. Jerrum and A. Sinclair. Conductance and the rapid mixing property for Markov chains: The approximation of the permanent resolved. In *Proceedings of the Twentieth ACM Symposium on Theory of Computing*, pages 235-244, 1988.

9. R. M. Karp and M. Luby. Monte Carlo algorithms for enumeration and reliability problems. In *Proceedings of the Twenty-fourth IEEE Symposium on Foundations of Computer Science*, 1983.

10. S. L. Lauritzen and D. J. Spiegelhalter. Local computations with probabilities on graphical structures and their application to expert systems. *Journal of the Royal Statistical Society* B 50 (19):157-224, 1988.

11. J. K. Patel and C. B. Read. *Handbook of the Normal Distribution*. Marcel Dekker, New York, NY, 1982.

12. J. Pearl. Fusion, propagation, and structuring in belief networks. *Artificial Intelligence*, 2:241-288, 1986.

13. J. Pearl. Evidential reasoning using stochastic simulation of causal models. *Artificial Intelligence*, 32:245-257, 1987.

14. J. Pearl. Addendum: Evidential reasoning using stochastic simulation of causal models. *Artificial Intelligence*, 33:131, 1987.